\documentclass{article}

\usepackage{microtype}
\usepackage{graphicx}
\usepackage{subcaption}
\usepackage{booktabs} 

\usepackage[hyphens]{url}
\usepackage{hyperref}

\usepackage[preprint]{icml2026}

\usepackage{amsmath}
\usepackage{amssymb}
\usepackage{mathtools}
\usepackage{amsthm}
\usepackage{soul}
\setstcolor{red}

\usepackage[capitalize,noabbrev]{cleveref}

\theoremstyle{plain}
\newtheorem{theorem}{Theorem}[section]
\newtheorem{proposition}[theorem]{Proposition}

\theoremstyle{definition}
\newtheorem{definition}[theorem]{Definition}

\theoremstyle{remark}

\newtheorem{example}{Example}

\newcommand{\Comment}[1]{\{#1\}}

\usepackage{tikz}
\usetikzlibrary{arrows.meta,positioning,calc}

\usepackage{multirow}

\usepackage[disable,textsize=tiny]{todonotes}

\newcommand{\Var}{\mathcal{V}}
\newcommand{\Lit}{\mathcal{L}}

\newcommand{\True}{\mathsf{true}}
\newcommand{\False}{\mathsf{false}}

\newcommand{\Annd}{\wedge}

\newcommand{\clause}[2]{\ensuremath{(#1 \lor #2)}}
\newcommand{\impl}[2]{\ensuremath{#1 \rightarrow #2}}
\newcommand{\implclause}[2]{\ensuremath{(\neg #1 \lor #2)}} %

\icmltitlerunning{Evaluating Robustness of Reasoning Models on Parameterized Logical Problems}

\newcommand*\circled[1]{\tikz[baseline=(char.base)]{
    \node[shape=circle,draw,inner sep=2pt] (char) {#1};}}

\begin{document}

\twocolumn[
  \icmltitle{Evaluating Robustness of Reasoning Models on Parameterized Logical Problems}

  \icmlsetsymbol{equal}{*}

  \begin{icmlauthorlist}
    \icmlauthor{Naïm Es-sebbani}{yyy}
    \icmlauthor{Esteban Marquer}{comp}
    \icmlauthor{Yakoub Salhi}{yyy}
    \icmlauthor{Zied Bouraoui}{yyy}
  \end{icmlauthorlist}

  \icmlaffiliation{yyy}{CRIL UMR 8188, Univ Artois, CNRS, France }
  \icmlaffiliation{comp}{GREYC, Université de Caen Basse Normandie, France}

  \icmlcorrespondingauthor{Naïm Es-sebbani}{essebbani@cril.fr}
  \icmlcorrespondingauthor{Esteban Marquer}{esteban.marquer@unicaen.fr}

  \icmlkeywords{Machine Learning, ICML}

  \vskip 0.3in
]



\printAffiliationsAndNotice{}  

\begin{abstract}
Logic provides a controlled testbed for evaluating LLM-based reasoners, yet standard SAT-style benchmarks often conflate surface difficulty (length, wording, clause order) with the structural phenomena that actually determine satisfiability. We introduce a diagnostic benchmark for \emph{2-SAT} built from parameterized families of structured 2--CNF formulas, where satisfiability is characterized by the implication graph and can be tuned along interpretable axes. Our generators isolate distinct competencies and failure modes: (i) contradiction-cycle UNSAT cores with controllable size and imbalance, (ii) SAT instances with a prescribed fraction of free variables to control solution multiplicity, (iii) planted backbones that modulate propagation, (iv) late bridge clauses that couple otherwise monotone regions to probe sensitivity to ordering and revision, and (v) symmetry/duplication variants that test abstraction under renaming and redundant structure. We evaluate LLM-based reasoners on decision accuracy and assignment validity, and quantify robustness under semantics-preserving perturbations such as clause reordering, filler clauses, and variable renaming. Across models, we observe sharp performance transitions under targeted structural interventions even when surface statistics are held fixed, revealing brittleness regimes that are invisible to aggregate SAT accuracy.
\end{abstract}

\section{Introduction}


Large reasoning models (LRMs) augment standard language models with training- and inference-time mechanisms that encourage explicit multi-step computation, and they have achieved strong empirical performance on mathematical reasoning and code generation \cite{DBLP:conf/iclr/PatelLRCRC23,bubeck2023sparksartificialgeneralintelligence}. Motivated by these gains, recent benchmarks increasingly emphasize combinatorial search and constraint satisfaction, including planning  and propositional satisfiability \cite{DBLP:journals/tmlr/ValmeekamSGK25}. While LRMs often outperform conventional LLMs in these settings, both model families frequently exhibit abrupt degradation once instance complexity crosses a threshold. One hypothesis is that LRMs approximate solver-like procedures through search-like behaviors such as self-correction and candidate selection \cite{hazra2025have}; if so, performance should be stable under semantics-preserving transformations and should degrade primarily with increases in structural difficulty.

The central question is whether such behaviors reflect transferable algorithmic competence or task-specific pattern reuse. A growing body of evidence points to brittle generalization: models can fail even when instances admit short solutions \cite{DBLP:conf/acl/AriyaniB0S25}, exhibit systematic errors on certain relational reasoning classes \cite{DBLP:journals/corr/abs-2510-23532}, and transfer poorly across closely related maze distributions \cite{DBLP:journals/corr/abs-2505-13775}. Prior work also suggests that performance can depend on producing traces with the “right” surface structure even when intermediate steps are not semantically faithful \cite{DBLP:journals/corr/abs-2502-07374}. These motivate evaluations that isolate genuine structure tracking from reliance on memorized trace templates and surface cues.

We advocate a more diagnostic evaluation regime based on parameterized problem families, where difficulty varies along explicit structural axes while the verification objective remains unchanged. Concretely, we focus on satisfiability of 2-CNF formulas (2-SAT). Although 2-SAT is polynomial-time solvable, it supports rich structural phenomena and admits an exact characterization via strongly connected components of the implication graph, enabling clean, verifiable perturbations. We design generators that target distinct competencies, including long-range implication tracking, localization of contradiction cores, and invariance to clause order and variable renaming. This fine-grained control enables stress tests beyond aggregate SAT/UNSAT accuracy and exposes brittleness regimes that are not visible under a single instance distribution.
Our contributions are threefold. (i) We introduce a diagnostic 2-SAT benchmark built from parameterized families of structured 2-CNF formulas with controllable implication-graph signatures and semantics-preserving perturbations (e.g., clause reordering, filler clauses, and variable renaming). (ii) We propose an evaluation suite that scores both decision accuracy and assignment validity, and quantifies robustness under targeted interventions. (iii) We present a systematic empirical study of LLM-based reasoners across these families, revealing sharp performance transitions and generator-specific brittleness even when instance size is held fixed.

\section{Related work}
\noindent\textbf{Reasoning in LLMs and LRMs.}
A large literature evaluates language models as reasoners across inference regimes, including deductive \cite{DBLP:conf/ijcai/ClarkTR20, saeed-etal-2021-rulebert}, abductive \cite{DBLP:conf/nips/KazemiYBKXIR23}, commonsense \cite{tian-etal-2021-diagnosing}, and logical or symbolic reasoning \cite{jiang-etal-2024-leanreasoner}. Beyond language-centric benchmarks, recent work probes models on algorithmic and combinatorial tasks that more directly instantiate search and constraint satisfaction, including graph search \cite{DBLP:journals/corr/abs-2510-22371}, planning \cite{DBLP:journals/tmlr/ValmeekamSGK25}, puzzles \cite{DBLP:journals/corr/abs-2506-06941}, propositional satisfiability \cite{hazra2025have}, and graph coloring \cite{DBLP:journals/corr/abs-2502-07087}. Across these settings, the required inference depth is often a strong predictor of difficulty \cite{parmar-etal-2024-logicbench}. Methodologically, three common approaches are used to elicit or support reasoning: prompting-based scaffolds such as CoT \cite{DBLP:conf/nips/KojimaGRMI22, DBLP:conf/iclr/Saparov023} and in-context exemplars \cite{wang-etal-2023-towards}; specialized reasoning models that improve over fine-tuning \cite{DBLP:journals/corr/abs-2409-13373}; and tool-oriented pipelines that translate problems into symbolic forms executed by external solvers or verifiers \cite{jiang-etal-2024-leanreasoner}.

\noindent\textbf{Logical benchmarks and SAT-based evaluation.}
A prominent line of work studies formal inference when logical statements are rendered in simplified natural language. \citet{DBLP:conf/ijcai/ClarkTR20} report near-perfect accuracy on Horn-rule entailment in such settings, but subsequent analyses find brittleness under distribution shift, consistent with shortcut learning rather than systematic rule-based generalization \cite{DBLP:conf/ijcai/ZhangLMCB23}. Related observations hold for satisfiability-style evaluation: \citet{DBLP:conf/aaai/0001S22} consider hard propositional SAT instances presented in simplified language and highlight sensitivity to training data selection and limited extrapolation to larger instances. Another approach seeks robustness by eliciting explicit proof traces, relying on the premise that reliable local steps can compose into longer derivations \cite{saha-etal-2020-prover, tafjord-etal-2021-proofwriter, DBLP:conf/iclr/CreswellSH23}.

\noindent\textbf{Logical reasoning datasets.}
Benchmarks such as LogiQA \cite{DBLP:conf/ijcai/LiuCLHWZ20}, ReClor \cite{DBLP:conf/iclr/YuJDF20}, BoardgameQA \cite{DBLP:journals/corr/abs-2306-07934}, and CLUTRR \cite{DBLP:conf/emnlp/SinhaSDPH19} probe logical and relational reasoning in text, but often entangle formal inference with world knowledge and linguistic priors. In contrast, datasets such as FOLIO \cite{DBLP:conf/emnlp/HanS0QRZCPQBSWS24}, RuleTaker \cite{DBLP:conf/ijcai/ClarkTR20}, and P-FOLIO \cite{DBLP:conf/emnlp/HanYSQRZQ0YLJZX24} more directly target knowledge-light logical inference. Logical puzzle corpora provide another controlled avenue \cite{DBLP:conf/emnlp/GiadikiaroglouL24}, including ZebraLogic \cite{DBLP:journals/corr/abs-2502-01100} and AutoLogi \cite{DBLP:journals/corr/abs-2502-16906}. Finally, some recent benchmarks evaluate models on SAT-style inputs more directly, e.g., PARAT \cite{DBLP:journals/corr/abs-2410-07432} and SATBench \cite{wei-etal-2025-satbench}, which frame satisfiability problems as natural language puzzles and leverage classical SAT solvers for verification.

\noindent\textbf{Position.}
Compared to prior logic and SAT benchmarks, we emphasize \emph{parameterized} instance families with fine-grained structural control and \emph{semantics-preserving perturbations} (e.g., clause order, filler clauses, and renaming invariances). This differs from SAT-in-natural-language settings that primarily vary surface realizations of fixed instances, and from proof/trace-based evaluations that require models to emit (and be graded on) explicit derivations. Instead, we keep the verification objective unchanged and probe whether models' decisions and assignments remain stable under transformations that preserve satisfiability while targeting specific structural competencies. Focusing on 2-SAT enables mechanistic instance design because satisfiability is characterized by the implication graph, allowing attribution of failures to concrete graph phenomena rather than to instance size or surface form alone.


\section{Problem setting}
\label{sec:problem}

We study satisfiability of 2--CNF formulas and design parameterized instance families whose difficulty can be tuned through interpretable structural properties. Throughout, let $\Var=\{x_1,\dots,x_n\}$ be a finite set of Boolean variables, and let $\Lit=\Var \cup \{\neg x \mid x\in\Var\}$ be its associated set of literals. Given a literal $\ell$, we write $\overline{\ell}$ for its complementary literal. We use the usual definitions $\ell\rightarrow \ell' := \overline{\ell}\vee \ell'$ and $(\ell \leftrightarrow \ell') := (\ell\rightarrow \ell')\wedge(\ell'\rightarrow \ell) \equiv \clause{\overline{\ell}}{\ell'}\wedge \clause{\overline{\ell'}}{\ell}$.

A \emph{2--CNF} is a conjunction of clauses, each clause a disjunction of at most two literals. 
A literal in a 2--CNF formula is \emph{pure} if it appears with only one polarity.
We write $n:=|\Var|$ for the number of variables and $m:=|\mathcal{C}|$ for the number of clauses. 

\noindent\textbf{Implication graph for 2-SAT.}
To decide satisfiability of a $2$--CNF formula, we associate with it an implication graph on vertex set $\Lit$. Each clause $\clause{\ell_1}{\ell_2}$ contributes the edges $\overline{\ell_1} \to \ell_2$ and $\overline{\ell_2} \to \ell_1$. We use $l\leadsto l'$ to denote that there is a path from $l$ to $l'$ in the implication graph.
A 2--CNF formula is unsatisfiable (UNSAT) iff there exists a variable $x$ such that $x$ and $\neg x$ belong to the same Strongly Connected Component (SCC) of the implication graph (equivalently, $x\leadsto \neg x$ and $\neg x\leadsto x$).
%

\noindent\textbf{Difficulty controls.} For each generator, we vary the instance structure and its presentation (clause order, variable renaming),
which does not change satisfiability but can stress heuristic sensitivity. For UNSAT generators, we mainly vary (i) the structure of the implication-graph witness of unsatisfiability (the
\emph{UNSAT core}) and (ii) the number of \emph{filler} clauses, i.e., clauses that can be satisfied
independently of the core and thus do not contribute to the contradiction. For SAT generators, we mainly vary (i) the number of \emph{critical variables}
(e.g., backbone variables), (ii) the total number of variables, and (iii) the
number of pure literals (literals that occur with only one polarity and hence do not constrain satisfiability).


\subsection{UNSAT Generator using Implication Cycles}

We pick distinct variables $v_1,\dots,v_L$ and for each we pick a literal
$\ell_i \in \{v_i, \neg v_i\}$ for $i=1,\dots,L$.
Then, we construct a directed cycle in the implication graph:
\begin{align*}
\textbf{Forward part:}&\qquad \ell_1 \to \ell_2 \to \cdots \to \ell_k \to \overline{\ell_1},\\
\textbf{Backward part:}&\qquad \overline{\ell_{1}} \to  \ell_{k+1}\to \cdots \to  \ell_L \to \ell_1.
\end{align*}
Encoding each implication $\impl{\ell}{\ell'}$ as the single 2-clause $\overline{\ell}\lor{\ell'}$ yields a core that forces both $\ell_1 \to \overline{\ell_1}$ and $\overline{\ell_1} \to \ell_1$, hence UNSAT.
The construction above contains one clause per displayed implication.
The forward part contributes $k$ clauses; the backward part contributes $(L-k+1)$ clauses, for a total of $L{+}1$ clauses.

\noindent\textbf{Parameters controlling difficulty.}
Difficulty is controlled by four parameters: (i) the \textbf{cycle split $(L,k)$}, where $L$ is the number of literals in the core cycle and $k$ is the position of the twist  to $\overline{\ell_1}$ (equivalently, the \textbf{imbalance} $k/L$, with extremes $k=1$ and $k=L$); (ii) the number of \textbf{filler clauses $r$} that are satisfiable independently of the core (controlling size/density); (iii) the \textbf{presentation}, i.e., whether core clauses are ordered along the implication chain or  shuffled and interleaved with fillers; and (iv) the \textbf{total number of variables $n\ge L$}, with variables outside the core appearing only in fillers (controlling the ratio $L/n$). The key structural difficulty is the imbalance $k/L$: when it is close to $0$ or $1$ (e.g., $k=1$ or $k=L$), one direction of the contradiction cycle is long while the other is short, forcing long-range implication tracking before closing the cycle.



\subsection{SAT Generator via Free Variables}

We choose a set $F\subseteq \Var$ of \emph{free variables} with $|F|=f$, meaning that the values of variables in $F$ can be chosen arbitrarily. For every remaining variable $y\in \Var\setminus F$, we choose a parent $p(y)\in F$ and a literal $\tau(y)\in\{p(y),\neg p(y)\}$, and enforce $y \leftrightarrow \tau(y)$.
Each such definition contributes two binary clauses (two implications). Every assignment to $F$ extends uniquely to a satisfying assignment on all of $\Var$.

\noindent\textbf{Parameters controlling difficulty.}
The number of variables $n$ and free variables $f$ determines the model count $2^f$.
The \textbf{free-to-bound ratio} $f/(n-f)$ controls coupling strength: smaller ratios induce stronger propagation from a small free set. Varying the \textbf{presentation order} by placing definition clauses early vs.\ late in the clause list to probe order sensitivity. The size of the formula is $2(n-f)$.
This isolates sensitivity to solution multiplicity (via $f$) and the amount of propagation required to produce a valid assignment.

\subsection{Backbone-based Method}
The goal is to construct a satisfiable $2$--CNF that is the conjunction of:
(i) clauses that infer a planted \emph{backbone} assignment, i.e., an assignment on a subset of variables that is forced in every satisfying assignment (these variables take the same truth value in all solutions), and
(ii) a remainder consisting only of monotone 2-clauses (all literals have the same polarity).
We choose a backbone $B\subseteq \mathcal V$ with a planted assignment
$b:B\to\{0,1\}$, and let $F=\mathcal V\setminus B$ be the free part.
We introduce a fresh auxiliary variable $a_x$ for each $x\in B$.
The formula is a conjunction of two parts:
\[
\Phi^\star  := \underbrace{\Phi_{\text{backbone}}}_{\text{infers }b\text{ using only 2-clauses}}
 \land\
\underbrace{\Phi_{\text{mono}}}_{\text{monotone 2-clauses over }F}.
\]

\noindent\textbf{(A) Backbone inference.}
For every $x\in B$, we add the following clauses using a fresh auxiliary $a_x$:
\[
\begin{aligned}
\text{Case } b(x)=1\!:&\quad (x \lor a_x) \land (x \lor \neg a_x),\\
\text{Case } b(x)=0\!:&\quad (\neg x \lor a_x) \land (\neg x \lor \neg a_x).
\end{aligned}
\]
Each pair is logically equivalent to the unit clause it emulates (respectively $x$ or $\neg x$), but uses only binary clauses; thus it fixes $x$ to $b(x)$ in all satisfying assignments.

\noindent\textbf{(B) Monotone remainder.}
We choose an orientation $\pi\in\{+,-\}$ and add only 2-clauses over variables in $F$:
\begin{align*}
\text{Case }\pi=-:\ &\text{add clauses }(\neg y \lor \neg z)\ \text{for }y,z\in F,\\
\text{Case }\pi=+:\ &\text{add clauses }(y \lor z)\ \text{for }y,z\in F.
\end{align*}
No variable from $B$ appears in $\Phi_{\text{mono}}$.
If $\pi=-$, setting all $y\in F$ to $0$ satisfies $\Phi_{\text{mono}}$; if $\pi=+$, setting all $y\in F$ to $1$ does. Together with the pinned backbone, $\Phi^\star$ is satisfiable.

\noindent\textbf{Parameters controlling difficulty.}
Let $|B|=\beta n$ and $m_{\text{mono}}$ be the number of monotone clauses added over $F$. The part  $\Phi_{\text{backbone}}$ has exactly $2|B|$ clauses and $|B|$ auxiliaries.
The total clause count is $|\mathcal{C}| = 2|B|+m_{\text{mono}}$.
Difficulty is tuned via the \emph{backbone fraction} $\beta$, the \emph{monotone density} relative to possible pairs $\frac{2 m_{\text{mono}}}{|F|^2}$, and the \textbf{orientation} $\pi$.



\subsection{Monotone Split with  Moving Cross-Polarity Bridge}
\label{sec:monotone-split-bridge}
We construct a $2$--CNF such that
(i) a positive-monotone part over variables $P$ (only clauses of the form $(p\lor p')$ with $p,p'\in P$),
(ii) a negative-monotone part over variables $N$ (only clauses of the form $(\neg q\lor \neg q')$ with $q,q'\in N$),
with $P\cap N=\varnothing$, and
(iii) a single bridge clause that couples both parts
$(\neg p^\star  \lor q^\star)$ with $p^\star\in P$ and  $q^\star\in N$.
The bridge is inserted at a chosen position in the clause list so we can probe when a solving method backtracks as it scans clauses.

\noindent\textbf{Parameters controlling difficulty.}
The number of clauses is $|\mathcal{C}| = m_+ + m_- + 1$ with $m_+$ and $m_-$ the number of positive and negative clauses, plus the bridge.
Densities $m_+/|P|$ and $m_-/|N|$ and the proportion of each part control how strongly each monotone part biases decisions.
\textbf{Bridge position} $s$ probes sensitivity to late coupling and potential revision during clause scanning; satisfiability is invariant under $s$.
In particular, we parametrize generation with the \textbf{overall variable density} $\frac{|P| + |N|}{2|\mathcal{C}|}$, the relative size of the positive part $m_+/|\mathcal{C}|$, and the position $s\in [1, |\mathcal{C}|-1]$ of the bridge among the clauses.

\subsection{Symmetry/Redundancy Probe}
\label{sec:concat-renamed-copy}
Given a base $2$--CNF  $\Psi$ over variables $\mathcal{V}$, we build a new instance
$\Phi := \Psi  \land \rho(\Psi)$,
where $\rho$ is a bijective renaming from $\mathcal{V}$ to a fresh, disjoint set $\mathcal{V}'$ (so $\mathcal{V}\cap \mathcal{V}'=\varnothing$).
Thus $\Phi$ is the concatenation of two isomorphic, independent copies of $\Psi$.
Comparing a model on $\Psi$ versus on $\Phi$ reveals the method's ability to avoid redoing symmetric/duplicated reasoning.

\noindent\textbf{Properties (independence and symmetry).} \emph{Independence}. $\mathcal{V}$ and $\mathcal{V}'$ are disjoint, so there is no logical interaction. \emph{Automorphism.} Swapping $\mathcal{V} \leftrightarrow \mathcal{V}'$ (i.e. applying $\rho$ or $\rho^{-1}$) leaves $\Phi$ invariant. $\Phi$ has a nontrivial symmetry.
\emph{Satisfiability.} $\Phi$ is SAT iff $\Psi$ is SAT. If $\Psi$ has $M$ models, then $\Phi$ has $M^2$ models (Cartesian product across copies). If $\Psi$ is UNSAT, so is $\Phi$.
Absent symmetry-awareness or abstraction that reuses proofs across variable renamings, a model may re-discover the same implications twice, once per copy.

\noindent\textbf{Variants and extensions.}
\emph{More copies.} We use $d>2$ disjoint renamings to form $\Psi\land \rho_2(\Psi)\land\cdots\land \rho_d(\Psi)$,
probing scaling under repeated symmetry.
\emph{Copy-wise shuffles.} Randomly permute the clause order within each copy; or apply a variable permutation per copy to avoid accidental alignment in indices.
\emph{Trigger placement.} If $\Psi$ has a late trigger clause that forces a contradiction, we place its renamed counterpart early (or vice versa) to study sensitivity across copies.


\label{sec:dataset}
\begin{table*}[t]
\footnotesize
    \centering
    \resizebox{0.9\textwidth}{!}{%

\begin{tabular}{ll|rr|rr|rr|rr|rr|rr|rr}
\toprule
 &  & \multicolumn{2}{c|}{Llama-3.3} & \multicolumn{2}{c|}{OLMo-3} & \multicolumn{2}{c|}{Phi4} & \multicolumn{2}{c|}{Phi4} & \multicolumn{2}{c|}{QwQ} & \multicolumn{2}{c|}{Qwen3-Next} & \multicolumn{2}{c}{GPT-OSS} \\
 & Model & \multicolumn{2}{c|}{70B-Instruct} & \multicolumn{2}{c|}{32B} & \multicolumn{2}{c|}{reasoning} & \multicolumn{2}{c|}{reasoning-plus} &  \multicolumn{2}{c|}{32B}& \multicolumn{2}{c|}{80B} & \multicolumn{2}{c}{120B} \\
Generator & $|\mathcal{C}|$ & Sat. & Wit. & Sat. & Wit. & Sat. & Wit. & Sat. & Wit. & Sat. & Wit. & Sat. & Wit. & Sat. & Wit. \\
\midrule
\multirow{7}{*}{ImplicationCycle} & 5 & 0.0 & --- & \textit{98.3} & --- & \textbf{83.3} & --- & 95.0 & --- & \textit{98.3} & --- & \textbf{100.0} & --- & \textit{91.7} & --- \\
 & 10 & 1.7 & --- & 95.0 & --- & \textbf{83.3} & --- & \textit{75.0} & --- & \textit{96.7} & --- & \textbf{98.3} & --- & 95.0 & --- \\
 & 15 & 15.0 & --- & \textbf{96.7} & --- & 61.7 & --- & 40.0 & --- & \textbf{68.3} & --- & \textbf{96.7} & --- & \textit{93.3} & --- \\
 & 20 & 18.3 & --- & \textbf{88.3} & --- & 41.7 & --- & 15.0 & --- & \textit{65.0} & --- & \textbf{96.7} & --- & \textbf{93.3} & --- \\
 & 50 & 11.7 & --- & 65.0 & --- & 3.3 & --- & 0.0 & --- & 40.0 & --- & \textbf{81.7} & --- & \textit{56.7} & --- \\
 & 75 & 13.3 & --- & \textit{50.0} & --- & 1.7 & --- & 0.0 & --- & 48.3 & --- & \textbf{73.3} & --- & 45.0 & --- \\
 & 100 & 20.0 & --- & \textit{51.7} & --- & 0.0 & --- & 0.0 & --- & 45.0 & --- & \textbf{76.7} & --- & 40.0 & --- \\
\midrule
\multirow{4}{*}{EquivalenceCore} & 10 & \textbf{100.0} & 1.7 & \textbf{100.0} & \textit{93.3} & \textbf{100.0} & 80.0 & \textbf{100.0} & 86.7 & \textbf{100.0} & 86.7 & \textbf{100.0} & 91.7 & \textbf{100.0} & \textbf{83.3} \\
 & 15 & \textit{98.3} & 0.0 & \textbf{100.0} & \textbf{55.0} & 90.0 & 35.0 & 95.0 & 35.0 & \textbf{96.7} & 45.0 & \textbf{100.0} & \textit{53.3} & \textbf{100.0} & 46.7 \\
 & 20 & 96.7 & 0.0 & 91.7 & 56.7 & 86.7 & \textit{21.7} & 83.3 & 38.3 & 86.7 & 43.3 & \textbf{100.0} & \textbf{63.3} & \textbf{100.0} & \textit{50.0} \\
 & 50 & \textit{93.3} & 0.0 & 83.3 & 1.7 & 76.7 & 1.7 & 68.3 & 0.0 & 83.3 & 1.7 & \textit{98.3} & \textbf{8.3} & \textbf{100.0} & \textbf{8.3} \\
\midrule
\multirow{4}{*}{Backbone} & 20 & \textbf{100.0} & 63.3 & \textbf{100.0} & 96.7 & \textbf{100.0} & 96.7 & \textbf{100.0} & \textbf{100.0} & \textbf{100.0} & \textbf{98.3} & \textbf{100.0} & 96.7 & \textbf{100.0} & \textit{96.7} \\
 & 50 & \textit{98.3} & \textit{71.7} & \textbf{98.3} & 83.3 & 98.3 & \textit{85.0} & 93.3 & 80.0 & 98.3 & 76.7 & \textbf{100.0} & 76.7 & \textbf{100.0} & \textit{98.3} \\
 & 75 & 96.7 & 51.7 & 90.0 & 36.7 & 96.7 & 58.3 & \textit{73.3} & 38.3 & 96.7 & 65.0 & \textbf{100.0} & 75.0 & 98.3 & \textbf{88.3} \\
 & 100 & \textbf{100.0} & 50.0 & 83.3 & 15.0 & 81.7 & 33.3 & \textbf{53.3} & 11.7 & 93.3 & 50.0 & 98.3 & 73.3 & \textbf{100.0} & \textbf{83.3} \\
\midrule
\multirow{7}{*}{MonoBridge} & 10 & \textbf{100.0} & 55.0 & \textbf{100.0} & 98.3 & \textbf{100.0} & 98.3 & \textbf{100.0} & 98.3 & \textbf{100.0} & 90.0 & \textbf{100.0} & \textbf{100.0} & \textbf{100.0} & \textbf{100.0} \\
 & 20 & \textbf{100.0} & 36.7 & \textbf{100.0} & 96.7 & \textbf{100.0} & 93.3 & \textbf{100.0} & 96.7 & 98.3 & 86.7 & \textbf{100.0} & \textbf{100.0} & \textbf{100.0} & \textbf{100.0} \\
 & 50 & 98.3 & 13.3 & 96.7 & 66.7 & 93.3 & 71.7 & 68.3 & 26.7 & 96.7 & 58.3 & \textbf{100.0} & 61.7 & \textbf{100.0} & 83.3 \\
 & 75 & \textbf{100.0} & 13.3 & 95.0 & 18.3 & 75.0 & 35.0 & 51.7 & 11.7 & 68.3 & 28.3 & 93.3 & \textit{60.0} & \textbf{100.0} & \textbf{68.3} \\
 & 100 & \textbf{100.0} & 1.7 & 88.3 & 1.7 & 80.0 & 20.0 & 48.3 & 0.0 & 83.3 & 13.3 & 91.7 & \textit{23.3} & \textbf{100.0} & \textit{53.3} \\
 & 150 & 96.7 & 3.3 & 80.0 & 1.7 & 83.3 & 1.7 & 58.3 & 0.0 & 78.3 & 1.7 & 93.3 & 21.7 & 90.0 & \textbf{31.7} \\
 & 200 & 98.3 & 0.0 & 66.7 & 0.0 & 78.3 & 0.0 & 65.0 & 0.0 & 83.3 & 0.0 & 93.3 & \textit{16.7} & 73.3 & \textit{23.3} \\
\bottomrule
\end{tabular}
}

    \caption{Satisfiability decision accuracy (Sat., \%) and witness validity (Wit., \%, SAT only). Each cell averages 10 formulas $\times$ 6 verbalizations. Difficulty at 0.5. Best in bold, second in italics.} 
    \label{tab:main-table}
\end{table*}

\section{Dataset construction}

\noindent\textbf{Verbalizers.}
We convert each 2--CNF into text using two approaches as illustrated in Fig \ref{fig:verbalisation}.
The \textbf{template verbalizer} deterministically renders each clause into a short natural-language statement, using a fixed mapping from variables to entity names within a theme. We use 6 different templates to reformulate the SAT problem into a more practical problem: a direct verbalization of how the clause reads (logic), letter passing between people (letter), assignment of people to a blue and a red team (team), affinities between invitees to constrain a social gathering (social), room lighting rules (room), door locking (door). Each template maps a clause to a fixed sentence pattern with named entities (e.g., $\lnot a \lor b$ becomes ``either a is on the Blue team or b is on the Red team''). The mapping is deterministic and invertible, so the logical content of each clause is preserved by construction.

The \textbf{LLM verbalizer} uses an LLM to rewrite each clause as a short narrative paragraph. We present the LLM with the two literals as natural-language statements derived from a thematic entity mapping (e.g., “the butler is innocent”, “the detective is the murderer”) and ask it to produce a paragraph expressing their disjunction. We use three themes (spy, heist, detective) to guide stylistic context.
To prevent information loss, each paragraph is checked by a two-step LLM validator: it (i) extracts the two referenced entities and (ii) infers whether each is stated positively or negatively. We compare the recovered clause to the original; mismatches trigger regeneration (up to three attempts), after which we fall back to a deterministic template. On the augmented dataset, 95.4\% of clauses validate on the first attempt and only 3.3\% use the fallback, indicating that the verbalizer is typically semantically faithful.

\noindent\textbf{Choice of generation parameters.}
To generate the dataset, we use difficulty the parameters described in \cref{sec:problem} for each generator. 
To facilitate sampling, each parameter is designed to be in $[0,1]$.
In our experiments, for each parameter, we split its possible values into thirds 
and generate data using an easy to manipulate value close to the center of each interval of the parameter space ($0.2\in[0;\frac{1}{3}]$, $0.5\in[\frac{1}{3},\frac{2}{3}]$, $0.8\in[\frac{2}{3},1]$).
When studying the impact of each parameter, we considered 4 additional values that correspond to the bounds of those intervals ($0, \frac{1}{3},\frac{2}{3}, 1$).
For generators where more than one major difficulty parameter identified, we vary only one parameter at a time and set the others to $0.5$ to minimize the number of settings to evaluate.
When generating data, the number of generated clauses is the main generation parameter, but minor differences might appear depending on the specifics of each generator. For instance, some generators are unable to generate an odd number of core clauses. Hereafter, number of clauses is given with a $\pm 1$ tolerance to best match the target difficulty.
\noindent\textbf{Pipeline.}
For a given generator parameter combination, we generate 10 formulas and produce, for each of them, either 6 verbalizations for the template verbalizer or 3 for the LLM verbalizer (only in the ablation study \cref{sec:ablation-verbalization}).


\section{Experiments}


Table~\ref{tab:main-table} reports results for seven public reasoning models (14B to 120B) using a fixed generator with all difficulty parameters set to 0.5, varying only the number of clauses. Model details appear in Appendix Table~\ref{tab:model-info}. For each setting, we sample 10 formulas and generate 6 template verbalizations per formula (60 prompts). We parse the SAT/UNSAT decision and, when SAT, an assignment, counting truncated or unparsable outputs as incorrect. Ablations focus on four models (Phi4-reasoning, Phi4-reasoning-plus, Qwen3-Next-80B, GPT-OSS-120B) under the same protocol unless noted.

\noindent\textbf{Number of clauses as the main difficulty parameter.}
%
Number of clauses as a primary control. Increasing $|\mathcal{C}|$ simultaneously increases (i) input length and parsing burden, and (ii) the amount of structure the model must integrate to reach a consistent global conclusion. We therefore treat $|\mathcal{C}|$ as the primary scaling axis. Importantly, $|\mathcal{C}|$ conflates core clauses (that witness SAT/UNSAT) with non-core clauses; in section \ref{sec:ablation-filler} we decouple these to test whether failures are driven by reasoning depth in the core or by distractor load. Phi4-reasoning-plus frequently produces outputs that exceed our maximum output budget, leading to truncated or unparsable answers even on moderate sizes. We report these cases as incorrect (per the protocol above). Details about the truncation rate is provided in Appendix~\cref{tab:truncation-rate}.

\noindent\textbf{Validation bias and variable assignment.}
%
Models can obtain deceptively high decision accuracy by over-predicting SAT. To separate decision priors from constructive reasoning, we additionally require a verifiable SAT witness: when predicting SAT, the model must output an assignment that satisfies all clauses under deterministic checking. We therefore report (i) SAT/UNSAT decision accuracy (Sat.), and (ii) witness validity (Wit.) on SAT instances. To quantify decision bias, we also report the model's SAT prediction rate on UNSAT instances (Appendix Table~\ref{tab:sat-bias}).
Each setting contains 10 independently generated formulas; the 6 verbalizations per formula are treated as repeated measurements and aggregated per formula for significance testing.


\begin{figure*}[t]
    \centering
    \includegraphics[height=0.2\textwidth]{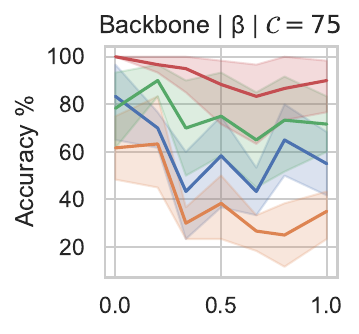}
    \hfill\hfill\hfill\hfill\hfill
    \includegraphics[height=0.2\textwidth]{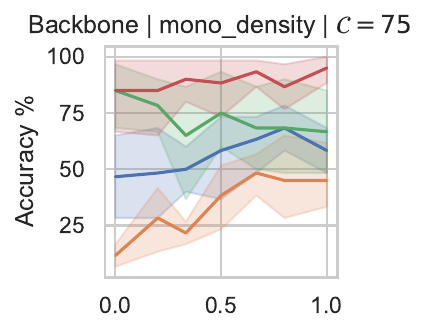}\hfill\hfill
    \includegraphics[height=0.2\textwidth]{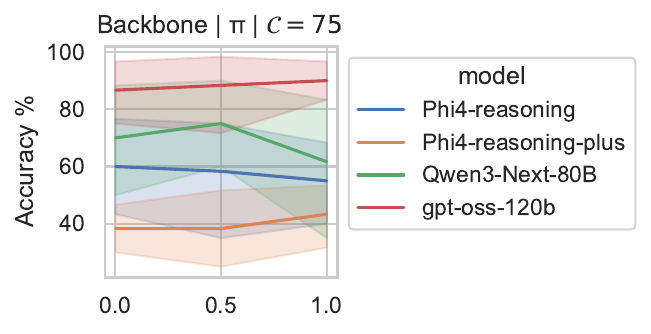}
    \caption{Main three difficulty parameters for Backbone at $|\mathcal{C}|=75$. Shade corresponds to 95\% confidence interval. For $\pi$, $\pi=0$ for a negative and $\pi=1$ for a positive monotonous part, while at $\pi=0.5$ the sign is randomly chosen.}
    \label{fig:backbone-parameters}
\end{figure*}
\begin{figure*}[t]
    \centering
    \includegraphics[height=0.2\textwidth]{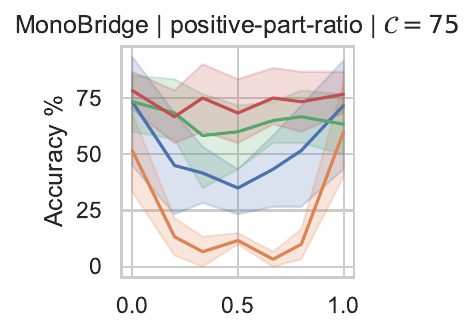}\hfill
    \includegraphics[height=0.2\textwidth]{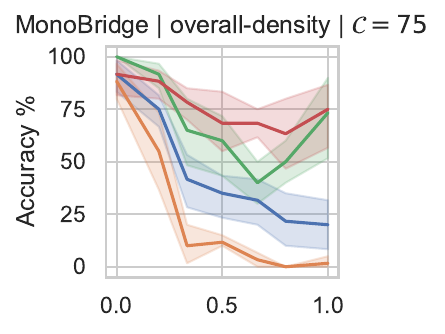}\hfill
    \includegraphics[height=0.2\textwidth]{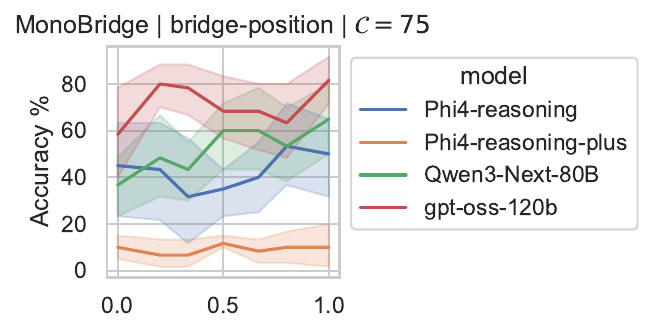}
    \caption{Main three difficulty parameters for MonoBridge at $|\mathcal{C}|=75$. Shade corresponds to 95\% confidence interval.}
    \label{fig:monobridge-positive-ratio}
\end{figure*}
\noindent\textbf{Generator difficulty at comparable formula size.}
With $|\mathcal{C}|$ and difficulty fixed (0.5), ImplicationCycle is consistently harder than the SAT generators: UNSAT requires finding a contradiction certificate (reachability $x\leadsto\neg x$ and $\neg x\leadsto x$) that depends on long-range implication integration, whereas EquivalenceCore and Backbone often allow local propagation over repeated equivalence patterns. Llama-3.3-70B-Instruct also shows a strong SAT bias, predicting SAT on 88.3\% of UNSAT instances (Appendix Table~\ref{tab:sat-bias}).
At $|\mathcal{C}|=50$, mean accuracy suggests ImplicationCycle and EquivalenceCore are hardest, then Backbone, with MonoBridge easiest. Analysis below isolate the impact of core size, distractors, and clause ordering.

\subsection{Impact of each difficulty parameters}
\label{sec:ablation-difficulty}

To assess whether generator parameters matter beyond $|\mathcal{C}|$, we sweep each parameter over a 7-point grid ${0, 0.2, 1/3, 0.5, 2/3, 0.8, 1}$ while fixing the others at 0.5. Each formula is one sample (10 per setting), with verbalizations aggregated per formula. We apply Friedman’s test across levels and Benjamini–Hochberg correction within each ablation family.
Sweeps are run at clause counts near each generator’s transition region: $|\mathcal{C}|=75$ for ImplicationCycle, MonoBridge, and Backbone, and $|\mathcal{C}|=20$ for EquivalenceCore. Appendix \cref{tab:significance} reports Friedman $p$-values at these sizes (7-point grid) and additional results aggregated across sizes and on a 3-point grid.

\noindent\textbf{MonoBridge parameters.}
The positive-part ratio has a significant effect ($p<0.05$ at $|\mathcal{C}|=75$): instances are hardest near a balanced split (ratio $\approx 0.5$) and easier when one monotone region dominates, consistent with a simple majority-assignment heuristic in the imbalanced regime and the need to reason about the cross-polarity bridge when balanced. Variable density also has a strong effect ($p\ll 0.001$): low density favors simplification via repeated variables, intermediate density produces a performance valley, and some large models recover at high density as the bridge becomes more salient. Qwen3\_Next-80B and GPT-OSS-120B in particular shift strategy around density $0.6$ to $0.8$, improving by exploiting the bridge rather than brute-force simplification. Bridge position is not significant, plausibly because models often treat clauses as an unordered constraint set rather than processing them sequentially.
%
%
The {bridge position} is not significant under our current protocol. The reason is that many models do not process clauses strictly sequentially: they often restate the problem as an unordered set of constraints before attempting simplification. 



%
\noindent\textbf{Backbone parameters.}
The three difficulty parameters identified for Backbone have limited impact on the performance compared to, for instance, the number of clauses.
First, the \textit{backbone fraction $\beta$} has the most significant effect on the performance ($p\text{-value} < 0.005$ for $|\mathcal{C}|=75$ and $p\text{-value} < 0.02$ overall), but this effect is marginal compared to the effect of the difficulty parameters of the other generators, with a drop of around 10\% in the average performance between $\beta=0$ and $\beta=1$. This indicates that the backbone part, that has pairs of related clauses, is more complex to handle than the monotonous part.
While the \textit{density of the monotonous part} (mono density) has the next most significant impact on performance, it is significant only when considering $|\mathcal{C}|=75$ where we observe an increase in performance of up to 20\% for the worst model ($p\text{-value} < 0.005$ for $|\mathcal{C}|=75$ and $p\text{-value} =0.31$ overall). Higher monotonous density means fewer distinct literals and stronger propagation.
Finally, the polarity $\pi$ of the monotonous part has no significant impact on the performance, with models manipulating literal polarity without issues in the CoT traces we analyzed (all $p\text{-values}> 0.9$). 









\noindent\textbf{EquivalenceCore parameter.}
\begin{figure}
    \centering
    \includegraphics[width=0.9\linewidth]{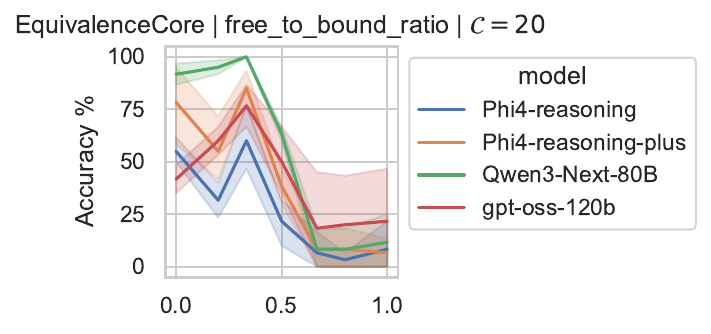}
    
    \caption{Main difficulty parameter for EquivalenceCore at $|\mathcal{C}|=20$. Shade corresponds to 95\% confidence interval.}
    \label{fig:equv-core-free-to-bound-ratio}
\end{figure}
The {free-to-bound variable ratio} has a very striking and consistent impact on performance as can be seen in \cref{fig:equv-core-free-to-bound-ratio}, with a consistent effect on all tested number of clauses 
($p\text{-value} < 0.005$ for $|\mathcal{C}|=20$ and $p\text{-value}=0.31$ overall).
High values of the ratio, corresponding to low propagation, are significantly harder to solve than low ratios, as the models rely on early simplification of equivalence classes through propagation. Across CoT traces, all models managed to properly identify equivalences from the $2$--CNF in its construction order. 



\noindent\textbf{ImplicationCycle parameter.} 
\begin{figure}
    \centering
    \includegraphics[width=0.8\linewidth]{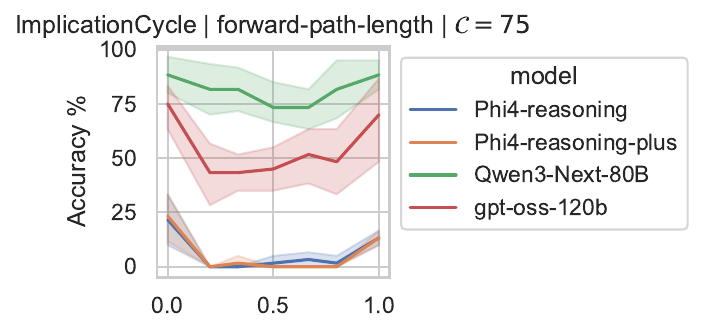}
    \caption{Main difficulty parameter for implication cycle at $|\mathcal{C}|=75$. Shade corresponds to 95\% confidence interval.}
    \label{fig:implication-cycle-split}
\end{figure}
The ImplicationCycle split parameter ($k/m$) affects performance mainly at larger sizes: extreme splits create one short side of the contradiction cycle, offering a shorter certificate path to track, while near-balanced splits force long-range reachability reasoning in both directions. The effect is significant at $|\mathcal{C}|=75$ but does not consistently generalize across sizes, so we see it as a targeted brittleness regime rather than a universal trend.





\subsection{Impact of non-core variables}
\label{sec:ablation-filler}

\begin{figure}
    \centering
    \includegraphics[width=.95\linewidth]{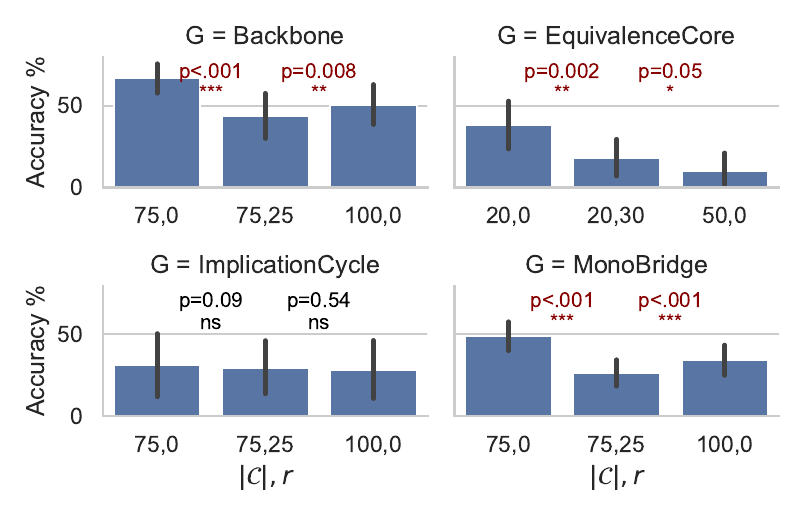}
    \caption{Comparison of decision accuracy on satisfiability (UNSAT problems) and SAT witness validity (SAT problems), between the construction order and a shuffling of the clauses. All difficulty parameters are set at $0.5$. Error bars are 95\% confidence intervals across models and verbalization templates.
    }
    \label{fig:ablation-filler}
\end{figure}

\label{sec:ablation-order}
\begin{figure}[t]
    \centering
    \includegraphics[width=\linewidth]{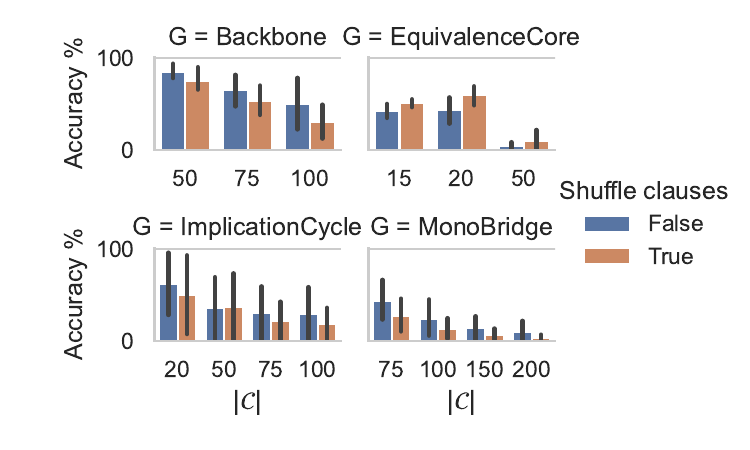}
    \caption{Comparison of accuracy between the construction order and a shuffling of the clauses. All difficulty parameters are set at $0.5$. Error bars are 95\% confidence intervals. For UNSAT problems, accuracy is measured on the prediction of satisfiability, and on variable assignment for SAT problems.}
    \label{fig:order-shuffle}
\end{figure}
To test sensitivity to variables irrelevant to satisfiability, we add \emph{filler} clauses over fresh variables. Let $V_{\text{fresh}}$ be disjoint from the core variable set and define literals $\ell_i\in{v_i,\neg v_i}$ for $v_i\in V_{\text{fresh}}$. We sample filler clauses from pairs of these literals, choosing $|V_{\text{fresh}}|$ so filler and core clauses have comparable variable density. We use two comparison regimes: fixed core size and matched total clause count. Concretely, we compare $(|C|=75,r=25)$ to $(75,0)$ and $(100,0)$, and $(|C|=20,r=30)$ to $(20,0)$ and $(50,0)$.
At difficulty 0.5, trends are visible but rarely significant, so we report results on the 3-level grid from \cref{sec:ablation-difficulty} (0.2, 0.5, 0.8). For all generators except ImplicationCycle, adjacent settings differ significantly (paired $t$-tests; see \cref{fig:ablation-filler}).
Filler clauses significantly hurt EquivalenceCore, Backbone, and MonoBridge when the core size is fixed. Under matched total clauses, EquivalenceCore behaves differently: a smaller equivalence core plus filler can be easier than a larger core alone. Trace inspection suggests models prioritize equivalences, but when they fail to compress equivalence classes, they repeatedly re-verify local consistency, making equivalences more costly than filler. For Backbone and MonoBridge, filler mainly adds noise, since their monotone regions already resemble filler structure, which further obscures the global pattern.


\begin{figure}[t]
    \centering
    \includegraphics[width=\linewidth]{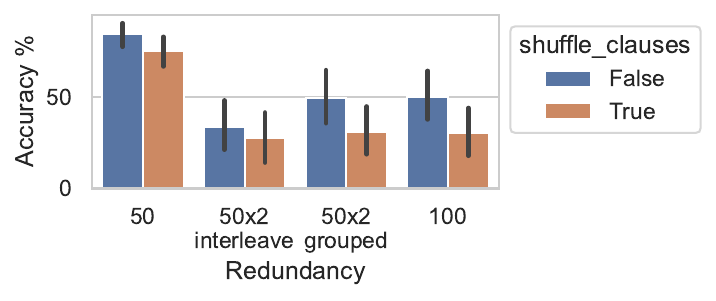}
    \caption{Comparison of accuracy when duplicating the clauses. All difficulty parameters are set at $0.5$. Error bars are 95\% confidence intervals. For UNSAT, accuracy is measured on the prediction of satisfiability, and on variable assignment for SAT problems.}
    \label{fig:ablation-repeat}
\end{figure}
\subsection{Sensitivity to clause order}


We compare in \cref{fig:order-shuffle} our four main models under two clause orders: \textit{(i)} the generator construction order (see \cref{sec:problem}) and \textit{(ii)} a random permutation. Clause order significantly affects performance for all generators (paired $t$-tests, $p<0.003$), with a weaker but still significant effect for ImplicationCycle ($p=0.035$). For EquivalenceCore, shuffling is beneficial. As discussed in \cref{sec:ablation-filler}, when implication structure is poorly formalized, contiguous implication chains can act as misleading local patterns; shuffling breaks these groupings by separating related clauses, reducing early commitment to an incorrect structure.

\subsection{Ability to handle repeated patterns}
\begin{figure}[t]
    \centering
    \includegraphics[width=\linewidth]{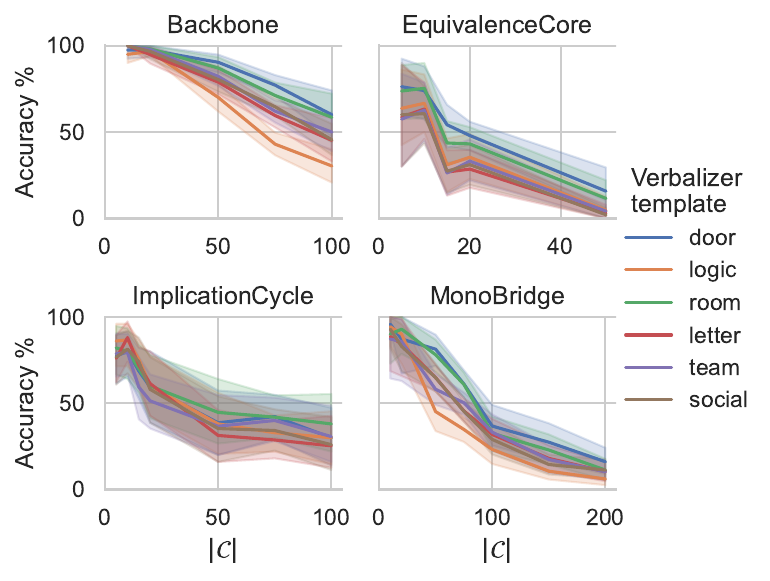}
    \caption{Decision accuracy on UNSAT and SAT witness validity against number of clauses for each generator, depending on the verbalization template used.
    Shade is the 95\% confidence interval.    }
    \label{fig:ablation-template-verbalizer}
\end{figure}
\begin{figure}[t]
    \centering
    \includegraphics[width=\linewidth]{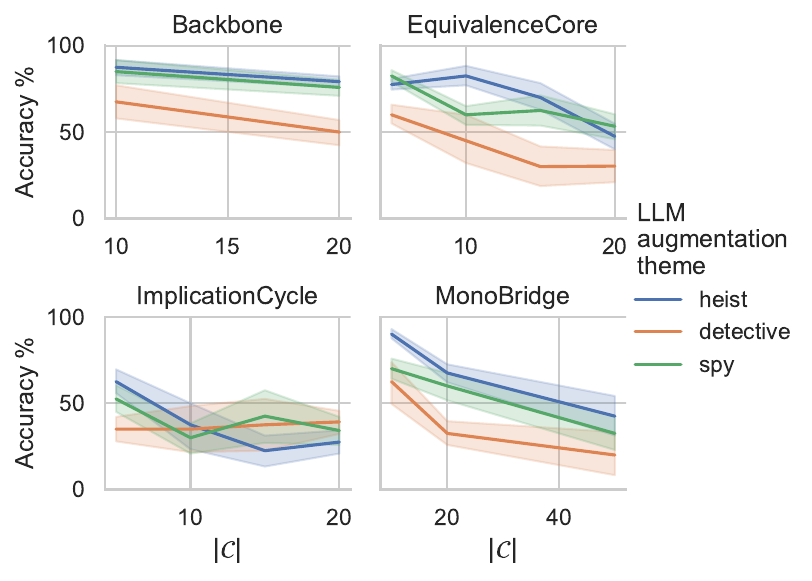}
    \caption{Decision accuracy on UNSAT  and SAT witness validity  against number of clauses for each generator, depending on the LLM verbalization theme used. Error bars are 95\% confidence intervals.     }
    \label{fig:ablation-llm-verbalizer}
\end{figure}

We report in \cref{fig:ablation-repeat} results for Backbone at medium difficulty (all parameters set to $0.5$) under clause repetition. We duplicate a base set of 50 clauses to obtain 100 clauses, using a fresh variable-to-entity mapping for the copy. We test two layouts: \textit{grouped}, where the copy forms the last 50 clauses, and \textit{interleave}, where copied clauses are randomly interleaved while preserving their internal order. We also compare against random clause permutations.
Performance at 50 clauses differs significantly from all other variants (independent $t$-tests, $p<0.001$). Shuffling significantly degrades performance within each setting (paired $t$-tests, $p<0.001$), except for 50$\times$2 \textit{interleave} ($p=0.036$). Differences between the 50$\times$2 and 100-distinct-clauses conditions are not significant ($p>0.1$), suggesting that repeating clauses is about as hard as adding distinct ones. Reasoning traces indicate that models rarely detect or exploit the repeated pattern.

\subsection{Verbalization strategy}\label{sec:ablation-verbalization}

As shown in \cref{fig:ablation-template-verbalizer}, template-only verbalizations yield similar performance across templates, indicating little sensitivity to surface wording when the clause structure is explicit. In contrast, LLM verbalizer substantially degrades accuracy (\cref{fig:ablation-llm-verbalizer}): at comparable clause sizes, performance drops by about 25 points relative to \cref{fig:ablation-template-verbalizer}. We attribute this mainly to increased contextual noise and weaker structural cues, which make clause extraction harder even though the underlying 2-CNF is information-preserving and can be recovered when models are explicitly prompted to do so (\cref{sec:dataset}).
Across themes, detective narratives are hardest, followed by spy, then heist, consistent with the intended stylistic bias. Detective stories tend to be more indirect, which further obscures the logical form. For ImplicationCycle, theme effects are smaller: the shared-literal chain induces more coherent narratives, improving verbalization consistency and partially mitigating theme differences. Overall, templates make the 2-CNF structure easy to recognize, whereas LLM narratives often impede structure recovery and lead models to miss global logical patterns.

\section{Conclusion}

We introduce a diagnostic 2-SAT benchmark based on parameterized 2-CNF families with controllable implication-graph structure, together with semantics-preserving perturbations such as clause reordering, variable renaming, and non-core fillers. This setup tests whether solver-like behavior remains stable when satisfiability is unchanged but the presentation and the required witness structure are altered.

Across several reasoning models, these targeted interventions induce sharp performance shifts even at fixed instance size. Long-cycle UNSAT instances are particularly brittle, and the gap between decision accuracy and assignment validity shows that correct predictions can mask weak witness construction. Sensitivity to ordering, fillers, and repetition further points to limited invariance and limited reuse of intermediate structure.

\newpage
\section*{Impact Statement}
This paper presents work whose goal is to advance the understanding of reasoning capabilities and failure modes in large language models through controlled diagnostic evaluation on logical satisfiability problems. We do not foresee specific negative societal consequences of this work. By exposing brittleness regimes that are invisible to aggregate accuracy metrics, our benchmark may contribute to more reliable deployment of LLM-based reasoners in safety-critical applications where logical consistency matters.


Looking ahead, two directions seem promising. First, extend the same recipe to richer constraint classes, for example k-SAT with planted structure, bounded-width CNF families, or SAT-to-proof settings, to test whether similar brittleness regimes persist. Second, use these diagnostics to guide model and decoding design toward explicit, reusable intermediate structure, such as induced implication graphs or SCC certificates, and toward invariances like order and renaming invariance that appear critical for robust reasoning.

\section*{Acknowledgment}
This work was supported by ANR-22-CE23-0002 ERIANA and was granted access to the HPC resources of IDRIS under the allocation 2026-AD011013338 made by GENCI.

\bibliography{references}
\bibliographystyle{icml2026}

\newpage
\appendix

\onecolumn
\section{Background}

Let $\mathcal{V}$ be a finite set of variables. 

A \emph{literal} is either a variable $p$ (positive) or its negation
$\neg p$ (negative).  Two literals are \emph{complementary} if one is
the negation of the other.

A \emph{clause} is a disjunction of literals.  
A  formula is in \emph{Conjunctive Normal Form} (CNF) if
it is a conjunction of clauses. A $2$--CNF formula is a CNF formula where each clause contains at most two literals.

An \emph{interpretation}   is a
function $\omega : \mathcal{V} \longrightarrow \{0,1\}$,
where $1$ stands for \emph{true} and $0$ for \emph{false}. It is extended to formulas as follows: $\omega(\neg \varphi)=1-\omega(\varphi)$, $\omega(\varphi\land\psi)= \min(\omega(\varphi),\omega(\psi))$, and $\omega(\varphi\lor\psi)= \max(\omega(\varphi),\omega(\psi))$.

A model of a formula $\varphi$ is an interpretation $\omega$ that satisfies the formula: $\omega(\varphi)=1$. In particular, $\omega$ is a model of a CNF formula $\varphi$ if and only if 
for each clause $c$ in $\varphi$, there exists a literal $l$ in $c$ such that $\omega(l)=1$.

A formula is \emph{satisfiable} if and only if it has at least one model; otherwise, it is \emph{unsatisfiable}.

\begin{definition}[Implication Graph]
Let $\varphi$ be a $2$--CNF formula. The {\em implication graph} of $\varphi$ is a directed graph $G=(V,E)$, where $V=Lit(Var(\varphi))$ and 
$E=\bigcup\{\{(\overline{l}, l'), (\overline{l'}, l)\}: l\vee l'\in\varphi\text{ or }(l\in\phi\text{ and }l'=l)\}$.
\end{definition}

We use $l\leadsto l'$ to denote that there is a path from $l$ to $l'$ in the implication graph.

\begin{definition}[Proof Cycle]
Let $\varphi$ be an unsatisfiable $2$--CNF formula. A {\em proof cycle} of the unsatisfiability of $\varphi$ is a cycle 
\circled{$~l~$}$\rightarrow$\circled{$l_1$}$\rightarrow\cdots{}$\circled{$l_{k}$}$\rightarrow$\circled{$~l~$} in the implication graph of $\phi$, where there exists $i\in\{1,\ldots{},k\}$ such that $l_i=\overline{l}$. 
\end{definition}

\begin{theorem}
A $2$--CNF formula is unsatisfiable iff it admits a proof cycle. 
\end{theorem}

\section{UNSAT Generator using Implication Cycles}

We create one clause per implication shown above:
$$
\boxed{
\begin{aligned}
&\implclause{\ell_1}{\ell_2}, \implclause{\ell_2}{\ell_3}, \ldots, \implclause{\ell_{k-1}}{\ell_k}, \implclause{\ell_k}{\neg \ell_1},\\
&\implclause{\neg \ell_1}{ \ell_{k+1}}, \implclause{ \ell_{k+1}}{ \ell_{k+2}}, \ldots, \implclause{ \ell_{m-1}}{\ell_m}, \implclause{ \ell_m}{\ell_1}.
\end{aligned}}
$$

We describe our generator in Algorithm~\ref{enerateUnsat2CNF}. 

\begin{example}[UNSAT core with $m{=}5$, $k{=}3$]
Let $(\ell_1,\ldots,\ell_5)=(v_1,\neg v_2,v_3, v_4, \neg v_5)$.  
Core clauses:
$$
\begin{aligned}
&(\neg v_1\lor \neg v_2), ( v_2\lor v_3), (\neg v_3\lor \neg v_1),\\
&( v_1\lor v_4), (\neg v_4\lor \neg v_5), ( v_5\lor v_1).
\end{aligned}
$$
This forces $v_1\leadsto \neg v_1$ and $\neg v_1\leadsto v_1$, hence UNSAT.
\end{example}

\begin{algorithm}[htbp]
\caption{\textsc{GenerateUnsat2CNF}$(n,m,k,r,\textit{shuffle})$}
\label{enerateUnsat2CNF}
\begin{algorithmic}[1]
\REQUIRE $n\ge m\ge 2$, $1\le k\le m$, $r\ge 0$, \textit{shuffle}$\in\{\True,\False\}$
\STATE Create variables $\Var=\{x_1,\dots,x_n\}$; pick distinct $v_1,\dots,v_m\in\Var$
\STATE For $i=1..m$, choose a polarity $\ell_i\in\{v_i,\neg v_i\}$
\STATE Initialize clause set $\mathcal{C}\gets\emptyset$
\Comment{Forward part}
\FOR{$i=1$ to $k-1$} \STATE $\mathcal{C}\gets \mathcal{C}\cup\{\implclause{\ell_i}{\ell_{i+1}}\}$ \ENDFOR
\STATE $\mathcal{C}\gets \mathcal{C}\cup\{\implclause{\ell_k}{\neg \ell_1}\}$
\Comment{Backward part}
\STATE $\mathcal{C}\gets \mathcal{C}\cup\{\implclause{\neg \ell_1}{\ell_{k+1}}\}$
\FOR{$i=k+1$ to $m-1$} \STATE $\mathcal{C}\gets \mathcal{C}\cup\{\implclause{\ell_i}{\ell_{i+1}}\}$ \ENDFOR
\STATE $\mathcal{C}\gets \mathcal{C}\cup\{\implclause{\ell_m}{\ell_1}\}$
\Comment{Fillers}
\FOR{$j=1$ to $r$}
  \STATE Add a filler clause $\clause{\lambda}{\lambda'}$ over $\Var$ (optionally avoid tautologies)
\ENDFOR
\IF{\textit{shuffle}} \STATE Randomly permute $\mathcal{C}$ \ENDIF
\STATE  Output $\Phi=\bigwedge_{C\in\mathcal{C}} C$ =0
\end{algorithmic}
\end{algorithm}

\section{SAT Generator via Free Variables}
For each $y\in \Var\setminus F$:
$$
\tau(y)=p(y):\quad \clause{\neg y}{p(y)} \Annd \clause{\neg p(y)}{y},
\qquad
\tau(y)=\neg p(y):\quad \clause{\neg y}{\neg p(y)} \Annd \clause{p(y)}{y}.
$$

This generator is described in Algorithm~\ref{GenerateSat2CNF}.

\begin{example}
Let $F=\{p\}$ (free) and define $q\leftrightarrow \neg p$, $r\leftrightarrow p$.
Clauses:
$$
\clause{\neg q}{\neg p} \Annd \clause{p}{q} \Annd 
\clause{\neg r}{p} \Annd \clause{\neg p}{r}.
$$
There are exactly two models, extending $p\in\{0,1\}$.
\end{example}

\begin{algorithm}[htbp]
\caption{\textsc{GenerateSat2CNF}$(n,f)$}
\label{GenerateSat2CNF}
\begin{algorithmic}[1]
\REQUIRE $n\ge 1$, $0\le f\le n$
\STATE Create variables $\Var=\{x_1,\ldots,x_n\}$; choose free set $F\subseteq \Var$ with $|F|=f$
\STATE $\mathcal{C}\gets \emptyset$
\FOR{each $y\in \Var\setminus F$}
  \STATE Pick $p(y)\in F$ and sign $s\in\{+1,-1\}$
  \IF{$s=+1$}
    \STATE $\mathcal{C}\gets \mathcal{C}\cup \{\clause{\neg y}{p(y)}, \clause{\neg p(y)}{y}\}$
  \ELSE
    \STATE $\mathcal{C}\gets \mathcal{C}\cup \{\clause{\neg y}{\neg p(y)}, \clause{p(y)}{y}\}$
  \ENDIF
\ENDFOR
\STATE Output $\Phi=\bigwedge_{C\in\mathcal{C}} C$ =0
\end{algorithmic}
\end{algorithm}

~\\
~\\

\section{Backbone-based Method}

\paragraph{Pseudocode:}
\begin{enumerate}
\item Input: $n$, $B\subseteq\mathcal V$, planted $b$, orientation $\pi\in\{+,-\}$, $m_{\text{mono}}$.
\item $F\leftarrow \mathcal V\setminus B$; initialize clause multiset $\mathcal C\leftarrow\emptyset$.
\item For each $x\in B$: introduce $a_x$ and add
  $(x\lor a_x),(x\lor\neg a_x)$ if $b(x)=1$, else $(\neg x\lor a_x),(\neg x\lor\neg a_x)$.
\item Repeat $m_{\text{mono}}$ times: sample $y,z\in F$, $y\neq z$, and add
  $(\neg y\lor\neg z)$ if $\pi=-$, else $(y\lor z)$.
\item Output $\Phi^\star=\bigwedge_{C\in\mathcal C} C$.
\end{enumerate}

\section{Monotone Split with a Moving Cross-Polarity Bridge}
\paragraph{Construction:}
\begin{enumerate}
  \item \textbf{Disjoint variable sets.} Partition the variables into $P=\{p_1,\dots,p_{n_+}\}$ and $N=\{n_1,\dots,n_{n_-}\}$ with $P\cap N=\emptyset$.
  \item \textbf{Positive-monotone core $\Phi_+$.} Generate $m_+$ clauses, each $(p_i\lor p_j)$ for some $p_i,p_j\in P$ (avoid tautologies and duplicates if desired).
  \item \textbf{Negative-monotone core $\Phi_-$.} Generate $m_-$ clauses, each $(\neg n_i\lor \neg n_j)$ for some $n_i,n_j\in N$.
  \item \textbf{Bridge clause $B$.} Pick endpoints $p^\star\in P$ and $n^\star\in N$ and set
        $$
        B := (\neg p^\star \lor n^\star).
        $$
  \item \textbf{Ordering (moving the bridge).} Fix an overall ordering of $\Phi_+\cup\Phi_-$ (e.g., all $\Phi_+$ then all $\Phi_-$ or any shuffle). 
        Let $M=m_++m_-$ be the number of monotone clauses.
        Choose an index $s\in\{1,\dots,M{+}1\}$ and \emph{insert $B$ as the $s$-th clause}.
        Varying $s$ across instances moves where the solver first encounters the coupling.
\end{enumerate}

\begin{proposition}[Satisfiable by construction]
The formula $\Phi:=\Phi_+\land\Phi_-\land B$ is satisfiable. 
\end{proposition}
For example, the assignment that 
$\text{set all }P\text{ to }1 \text{ and set }n^\star\text{ to }1\text{ and all other }N\setminus\{n^\star\}\text{ to }0
$
satisfies $\Phi_+$ (positives are $1$), satisfies $B$ (since $n^\star=1$), and satisfies every $(\neg n_i\lor \neg n_j)$ in $\Phi_-$ (at least one of $n_i,n_j$ is $0$).

\noindent
\textbf{Backtracking trigger.}
If a solver's early decisions tend to make $p^\star=1$ (natural for $\Phi_+$) and $n^\star=0$ (natural for $\Phi_-$), then when $B$ is finally processed it becomes $(\neg 1 \lor 0)$ and conflicts, forcing a revision (unit propagation or backtrack). By changing the insertion index $s$, we control when this conflict becomes visible.

\subsection*{Pseudocode}
\begin{enumerate}
  \item Input: $n_+,n_-,m_+,m_-,p^\star\in P,n^\star\in N$, insertion index $s\in[1,M{+}1]$.
  \item Build lists $C_+$ of $m_+$ clauses $(p\lor p')$ over $P$ and $C_-$ of $m_-$ clauses $(\neg n\lor \neg n')$ over $N$.
  \item Concatenate $C:=\text{order}(C_+,C_-)$ (any fixed order or shuffle); let $M=|C|$.
  \item Insert $B=(\neg p^\star\lor n^\star)$ at position $s$ in $C$.
  \item Output $\Phi=\bigwedge_{i=1}^{M+1} C[i]$.
\end{enumerate}

\section{Symmetry/Redundancy Probe}
\subsection*{Construction}
\begin{enumerate}
  \item \textbf{Base formula.} Let $\Psi=\bigwedge_{c\in \mathcal{C}} c$ be any $2$--CNF over variables $V=\{x_1,\dots,x_n\}$.
  \item \textbf{Disjoint renaming.} Choose a bijection $\rho:V\to V'=\{x'_1,\dots,x'_n\}$ with $V\cap V'=\varnothing$.
        Extend $\rho$ to literals and clauses in the obvious way (e.g., $\rho(\neg x)=\neg \rho(x)$, $\rho(\ell_1\lor \ell_2)=\rho(\ell_1)\lor \rho(\ell_2)$).
  \item \textbf{Concatenation.} Define
        $$
        \Phi := \Psi  \land \rho(\Psi).
        $$
        Optionally, {order} the clauses as (i) all of $\Psi$ then all of $\rho(\Psi)$ ({grouped}),
        or (ii) an {interleaving} of the two clause lists (to blur copy boundaries).
\end{enumerate}

\section{Verbalization Templates and Prompts}\label{app:templates}

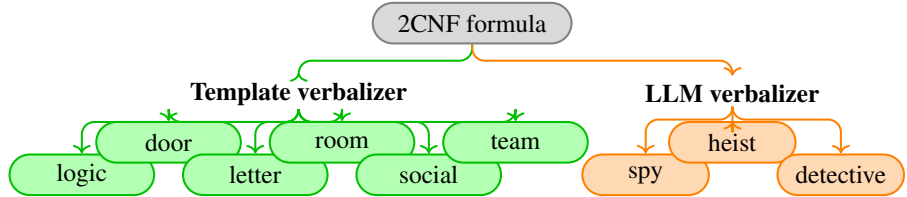
\begin{figure}[t]
    \centering
    \resizebox{.7\textwidth}{!}{%
\newcommand{\tinystubs}[1]{
  \node[tiny stub] (#1-stub2) at ($(#1.south)+(0,-1.25)$) {};
  \node[tiny stub] (#1-stub1) at ($(#1-stub2)+(0.25,0)$) {};
  \node[tiny stub] (#1-stub3) at ($(#1-stub2)+(-0.25,0)$) {};
  \draw[dashed, ->, draw=orange!50] (#1) -- (#1-stub2.north);
  \draw[dashed, ->, draw=orange!50] (#1) -- (#1-stub1.north);
  \draw[dashed, ->, draw=orange!50] (#1) -- (#1-stub3.north);
}%
\begin{tikzpicture}[
  pill/.style={rectangle, rounded corners=8pt, draw, thick, minimum height=16pt, align=center},
  grey pill/.style={pill, fill=gray!30, draw=gray, minimum width=2.75cm},
  green pill/.style={pill, fill=green!30, draw=green!75!black, minimum width=2cm},
  orange pill/.style={pill, fill=orange!30, minimum width=1.75cm, draw=orange},
  fan arrow green/.style={thick, ->, draw=green!75!black, rounded corners=5pt},
  fan arrow orange/.style={thick, ->, draw=orange, rounded corners=5pt},
  branch title/.style={font=\bfseries, align=center},
  yscale=.75,
  xscale=1.2
]

\node[grey pill] (formula) at (4.5, 0) {2CNF formula};

\node[branch title] (tpl-title) at (2.5, -1.3) {Template verbalizer};
\node[branch title] (llm-title) at (7.5, -1.3) {LLM verbalizer};

\draw[fan arrow green] (formula.south) -- ++(0,-0.3) -| (tpl-title.north);
\draw[fan arrow orange] (formula.south) -- ++(0,-0.3) -| (llm-title.north);

\node[green pill] (logic)  at (0, -2.8) {logic};
\node[green pill] (door)   at (1, -2.2) {door};
\node[green pill] (letter) at (2, -2.8) {letter};
\node[green pill] (room)   at (3, -2.2) {room};
\node[green pill] (social) at (4, -2.8) {social};
\node[green pill] (team)   at (5, -2.2) {team};

\draw[fan arrow green] (tpl-title.south) -- ++(0,-0.15) -| (logic.north);
\draw[fan arrow green] (tpl-title.south) -- ++(0,-0.15) -| (door.north);
\draw[fan arrow green] (tpl-title.south) -- ++(0,-0.15) -| (letter.north);
\draw[fan arrow green] (tpl-title.south) -- ++(0,-0.15) -| (room.north);
\draw[fan arrow green] (tpl-title.south) -- ++(0,-0.15) -| (social.north);
\draw[fan arrow green] (tpl-title.south) -- ++(0,-0.15) -| (team.north);

\node[orange pill] (spy)       at (6.5, -2.8) {spy};
\node[orange pill] (heist)     at (7.5, -2.2) {heist};
\node[orange pill] (detective) at (8.75,-2.8) {detective};

\draw[fan arrow orange] (llm-title.south) -- ++(0,-0.15) -| (spy.north);
\draw[fan arrow orange] (llm-title.south) -- ++(0,-0.15) -| (heist.north);
\draw[fan arrow orange] (llm-title.south) -- ++(0,-0.15) -| (detective.north);

\end{tikzpicture}
}
\caption{Summary of the verbalization process}
    \label{fig:verbalisation}
\end{figure}

\subsection{Template Verbalizers}

Each template verbalizer converts a 2-CNF clause $(\ell_1 \lor \ell_2)$ into a natural language sentence following a fixed pattern. The verbalizer also provides a context preamble and a final yes/no question. \Cref{tab:verbalizer-summary} summarizes the six schemes.

\begin{table*}[htbp]
\centering
\small
\begin{tabular}{llllp{5cm}}
\toprule
\textbf{Scheme} & \textbf{True value} & \textbf{False value} & \textbf{Entity naming} & \textbf{Context preamble} \\
\midrule
\texttt{logic}  & true & false & Letters: A, B, \ldots, Z, AA, AB, \ldots & Consider these logical relationships: \\
\texttt{letter} & has & doesn't have & Person names & Consider these rules about who has the letter: \\
\texttt{team}   & Red & Blue & Person names & We have team assignment rules: \\
\texttt{social} & attends & doesn't attend & Person names & You're planning a party with these attendance conditions: \\
\texttt{room}   & lit & dark & room 1, room 2, \ldots & A building has lighting rules: \\
\texttt{door}   & open & closed & door 1, door 2, \ldots & A facility has door coupling rules: \\
\bottomrule
\end{tabular}
\caption{Summary of the six template verbalizers. Each row shows the scheme name, the context preamble, the true/false value labels, and the entity naming convention.}
\label{tab:verbalizer-summary}
\end{table*}

\paragraph{Clause-to-text patterns.}
Below are the exact sentence templates used for each scheme. In all cases, $A$ and $B$ denote the entity names of the two literals' variables, and the polarity determines which value (true or false) is used.

\begin{itemize}
\item \textbf{logic:} \verb|Either {A} is true or {B} is false (or both).|
\item \textbf{letter:} \verb|Either {A} has the letter or {B} doesn't have the letter (or both).|
\item \textbf{team:} \verb|Either {A} is on the Red team or {B} is on the Blue team.|
\item \textbf{social:} \verb|{A} attends or {B} doesn't attend (or both).|
\item \textbf{room:} \verb|room {i} is lit or room {j} is dark (or both).|
\item \textbf{door:} \verb|door {i} is open or door {j} is closed (or both).|
\end{itemize}

\paragraph{Final questions.}
\begin{itemize}
\item \textbf{logic:} Can you assign truth values to all statements without creating a contradiction?
\item \textbf{letter:} Is there a consistent way for people to hold or not hold the letter?
\item \textbf{team:} Is there a valid team assignment that satisfies all constraints?
\item \textbf{social:} Can you create a guest list that respects everyone's conditions?
\item \textbf{room:} Can all the lighting rules be satisfied simultaneously?
\item \textbf{door:} Is there a configuration of doors that satisfies all the rules?
\end{itemize}

\paragraph{Assembled output.}
For each formula, the verbalizer concatenates the context preamble, one sentence per clause, a blank line, and the final question. For example, with the \texttt{logic} scheme and clauses $(\neg A \lor B)$, $(A \lor \neg C)$:

\begin{verbatim}
Consider these logical relationships:
Either A is false or B is true (or both).
Either A is true or C is false (or both).

Can you assign truth values to all statements
without creating a contradiction?
\end{verbatim}

\subsection{Evaluation Prompt}\label{app:eval-prompt}

The evaluation prompt is appended after the verbalized formula. Two variants exist depending on whether template or LLM verbalization is used.

\paragraph{Template verbalizer prompt.}
\begin{verbatim}
{verbalized formula text}

Think step by step.

If a valid assignment exists, provide it as JSON
using "{true_val}" or "{false_val}" as values.
Example format:
{
  "{entity_1}": "{true_val}",
  "{entity_2}": "{false_val}",
  "{entity_3}": "{true_val}"
}

End your response with exactly one of:
- "The answer is: Yes" (if a valid assignment exists)
- "The answer is: No" (if no valid assignment exists)
\end{verbatim}
where \texttt{\{true\_val\}} and \texttt{\{false\_val\}} are the scheme-specific values from \cref{tab:verbalizer-summary}, and the example JSON shows the first three entities.
\newpage 
\paragraph{LLM verbalizer prompt.}
\begin{verbatim}
{augmented story text}

Think step by step.

If a valid assignment exists, provide it as JSON
with one of the allowed values for each entity.

Entities and their possible values:
  - "{entity_name}": "{true_desc}" or "{false_desc}"
  ...

Example JSON format:
{
  "entity_name": "value"
}

End your response with exactly one of:
- "The answer is: Yes" (if a valid assignment exists)
- "The answer is: No" (if no valid assignment exists)
\end{verbatim}
where each entity is listed with its theme-specific true/false descriptions.

\subsection{LLM Story Generation Prompts}

The LLM verbalizer uses a multi-turn conversation to generate narrative clues. Below are the exact prompts used at each step.

\paragraph{Step 1: Global context generation.}
\begin{verbatim}
You are a mystery story writer. Create a short
introduction for a detective story.

CHARACTERS/OBJECTS in this story:
- {entity_1_name}
- {entity_2_name}
...

Requirements:
- Set the scene for a mystery with clues to discover
- Briefly introduce each character/object naturally
  in the narrative
- Do NOT mention any specific rules or relationships
  yet
- Keep it brief and atmospheric

Write only the introduction paragraph.
\end{verbatim}

\newpage 
\paragraph{Step 2: Per-clause narrative generation.}
For each clause $(\ell_1 \lor \ell_2)$, the following prompt is appended to the conversation:
\begin{verbatim}
Write clue {i}/{total} for our mystery story.

This clue must express that AT LEAST ONE of these
statements is true:
- Statement A: {entity_1_name} {state_1}
- Statement B: {entity_2_name} {state_2}

(It's possible both are true, but at least one
MUST be true.)

Write 2-3 sentences that naturally convey this
logical rule through detective reasoning, witness
testimony, or evidence analysis. Be creative but
ensure the logic is clear.

Write ONLY the clue paragraph:
\end{verbatim}

\paragraph{Step 3a: Entity extraction (validation).}
\begin{verbatim}
From this clue, identify the TWO main entities
being discussed.

CLUE: "{generated_paragraph}"

AVAILABLE ENTITIES:
["entity_1", "entity_2", ...]

Which two entities does this clue primarily discuss?
Respond with ONLY a JSON array of exactly 2 entity
names: ["entity1", "entity2"]
\end{verbatim}

\newpage 

\paragraph{Step 3b: State extraction (validation).}
For each of the two extracted entities:
\begin{verbatim}
In this clue, what is stated about {entity_name}?

CLUE: "{generated_paragraph}"

The clue says that {entity_name}:
A) {true_description}
B) {false_description}

Which state does the clue indicate?
Answer with ONLY "A" or "B".
\end{verbatim}

If validation fails after 3 attempts, a programmatic fallback sentence is used:

\verb|Either {entity_1} {state_1}, or {entity_2} {state_2} (or both).|

\section{Information about model used}
\cref{tab:model-info} reports general model information.
\begin{table}[t]
\centering
\begin{tabular}{lllrl}
\toprule
\textbf{Model} & \textbf{Model ID} & \textbf{Param.} & \textbf{Max Tokens} & \textbf{License} \\
\midrule
Phi 4 reasoning & microsoft/Phi-4-reasoning & 15B & 32K & MIT\\
Phi 4 reasoning plus & microsoft/Phi-4-reasoning-plus & 15B & 32K & MIT\\
Qwen3-Next 80B & Qwen/Qwen3-Next-80B-A3B-Thinking & 81B & 262K & Apache 2.0 \\
GPT-OSS-120B & openai/gpt-oss-120b & 120B & 131K & Apache 2.0 \\
OLMo-3 32B & allenai/Olmo-3.1-32B-Think & 32B & 66K & Apache 2.0 \\
QwQ 32B & Qwen/QwQ-32B & 33B & 41K & Apache 2.0  \\
Llama-3.3 70B-Instruct & meta-llama/Llama-3.3-70B-Instruct & 71B & 66K & Llama 3.3 \\
\bottomrule
\end{tabular}
\caption{Overview of the reasoning models.}
\label{tab:model-info}
\end{table}

\section{Truncation rate}
\cref{tab:truncation-rate} reports output truncation rates per model.
\begin{table}[t]
\centering                                                                                             
              
\begin{tabular}{lrr}                                                   
\toprule                                                          
\textbf{Model} & \textbf{Max Tokens} & \textbf{Truncation Rate (\%)} \\
\midrule       
Phi-4-reasoning         & 16 384 & 14.9 \\                                                                      
Phi-4-reasoning-plus    & 16 384 & 30.5 \\      
Qwen3-Next 80B          & 32 768 & 1.3 \\      
GPT-OSS-120B            & 32 768 & 0.0 \\                                       
OLMo-3 32B              & 32 768 & 3.0 \\      
QwQ 32B                 & 32 768 & 3.6 \\      
Llama-3.3 70B-Instruct  & 32 768 & 0.0 \\      

\bottomrule                                       
\end{tabular}                                      
\caption{Truncation rates by model under fixed output budget.}\label{tab:truncation-rate}                                   
\end{table}

\section{SAT Prediction Bias on UNSAT Instances}  

Table~\ref{tab:sat-bias} reports the proportion of UNSAT instances for which each model incorrectly predicted SAT. A high rate indicates a tendency to over-predict satisfiability regardless of the actual logical structure. Llama-3.3-70B-Instruct exhibits a particularly strong SAT bias (88.3\%), suggesting it defaults to predicting satisfiability rather than performing genuine constraint reasoning.
  
\begin{table}[t]                                                   
\centering         
                                      
\begin{tabular}{lr}                                                     
\toprule             
\textbf{Model} & \textbf{SAT Prediction (\%)} \\                                                     
\midrule                                    
Phi-4-reasoning & 33.3 \\                               
Phi-4-reasoning-plus & 20.2 \\                                     
Qwen3-Next 80B & 10.5 \\                                       
GPT-OSS-120B & 26.4 \\                                         
OLMo-3 32B & 21.7 \\                                            
QwQ 32B & 34.0 \\                             
Llama-3.3 70B-Instruct & 88.3 \\                                                  
\bottomrule                                  
\end{tabular}                                                 
\caption{SAT prediction rate on UNSAT instances (ImplicationCycle generator). Higher values indicate stronger bias toward predicting satisfiability.}\label{tab:sat-bias}                                                            
\end{table} 

\section{Statistical significance results}\label{sec:significance}
Significance results for Friedman’s test for repeated measurements across difficulty parameter levels with Benjamini–Hochberg correction are reported in \cref{tab:significance}.
\begin{table}
    \centering
    \begin{tabular}{rrrl}
\toprule
Generator & Difficulty parameter & $|\mathcal{C}|$ & $p$-value \\
\midrule
\multirow[t]{13}{*}{Backbone} & \multirow[t]{6}{*}{Monotonous density} & Overall & $3.363e^{-3}$ \\
 &  & 20 & $4.724e^{-1}$ \\
 &  & 50 & $3.813e^{-2}$ \\
 &  & 75 & $4.724e^{-1}$ \\
 &  & 75 7-point & $3.125e^{-1}$ \\
 &  & 100 & $3.877e^{-2}$ \\
\cline{2-4}
 & \multirow[t]{6}{*}{$\beta$} & Overall & $1.865e^{-2}$ \\
 &  & 20 & $1.462e^{-1}$ \\
 &  & 50 & $5.971e^{-2}$ \\
 &  & 75 & $3.877e^{-2}$ \\
 &  & 75 7-point & $4.310e^{-3}$ \\
 &  & 100 & $3.679e^{-1}$ \\
\cline{2-4}
 & $\pi$ & 75 7-point & $9.355e^{-1}$ \\
\midrule
\multirow[t]{5}{*}{EquivalenceCore} & \multirow[t]{5}{*}{Free-to-bound variable ratio} & Overall & $2.754e^{-5}$ \\
 &  & 15 & $1.832e^{-2}$ \\
 &  & 20 & $1.832e^{-2}$ \\
 &  & 20 7-point & $1.296e^{-3}$ \\
 &  & 50 & $8.208e^{-2}$ \\
\midrule
\multirow[t]{9}{*}{ImplicationCycle} & \multirow[t]{9}{*}{Forward path relative length} & Overall & $8.116e^{-1}$ \\
 &  & 5 & $7.643e^{-2}$ \\
 &  & 10 & $5.488e^{-1}$ \\
 &  & 15 & $3.679e^{-1}$ \\
 &  & 20 & $2.574e^{-1}$ \\
 &  & 50 & $4.412e^{-1}$ \\
 &  & 75 & $2.725e^{-1}$ \\
 &  & 75 7-point & $7.257e^{-3}$ \\
 &  & 100 & $5.292e^{-1}$ \\
\midrule
\multirow[t]{21}{*}{MonoBridge} & \multirow[t]{7}{*}{Bridge relative position} & Overall & $1.869e^{-1}$ \\
 &  & 50 & $1.889e^{-1}$ \\
 &  & 75 & $7.788e^{-1}$ \\
 &  & 75 7-point & $2.154e^{-1}$ \\
 &  & 100 & $1.738e^{-1}$ \\
 &  & 150 & $1.496e^{-1}$ \\
 &  & 200 & $1.353e^{-1}$ \\
\cline{2-4}
 & \multirow[t]{7}{*}{Overall variable density} & Overall & $2.073e^{-7}$ \\
 &  & 50 & $4.979e^{-2}$ \\
 &  & 75 & $1.832e^{-2}$ \\
 &  & 75 7-point & $2.719e^{-3}$ \\
 &  & 100 & $3.813e^{-2}$ \\
 &  & 150 & $9.697e^{-2}$ \\
 &  & 200 & $2.307e^{-2}$ \\
\cline{2-4}
 & \multirow[t]{7}{*}{Positive part ratio} & Overall & $1.158e^{-3}$ \\
 &  & 50 & $4.979e^{-2}$ \\
 &  & 75 & $4.724e^{-1}$ \\
 &  & 75 7-point & $2.944e^{-2}$ \\
 &  & 100 & $4.979e^{-2}$ \\
 &  & 150 & $4.412e^{-1}$ \\
 &  & 200 & $2.231e^{-1}$ \\
\midrule
\bottomrule
\end{tabular}
    \caption{Statistical significance results for each difficulty parameter}
    \label{tab:significance}
\end{table}

\end{document}